\documentclass{amia}

\usepackage[T1]{fontenc}
\usepackage[utf8]{inputenc}

\usepackage{graphicx}
\usepackage[labelfont=bf]{caption}
\usepackage[superscript,nomove]{cite}
\usepackage{url}

\usepackage{amsmath,amssymb}

\usepackage{booktabs}
\usepackage{multirow}

\usepackage{color}
\usepackage{soul}

\usepackage{microtype}

\usepackage{doi}
\usepackage{placeins}

\begin{document}
	
	\title{CTG-DB: An Ontology-Based Transformation of ClinicalTrials.gov to Enable Cross-Trial Drug Safety Analyses}
	
	\author{Jeffery L. Painter, MS, JD$^{1}$, François Haguinet, MS$^{2}$, Andrew Bate, PhD$^{3,4}$}
		
	\institutes{
		$^1$GSK, Durham, NC, USA; \\
		$^2$GSK, Wavre, Belgium; \\
		$^3$GSK, London, UK; \\
		$^4$London School of Hygiene \& Tropical Medicine, London, UK
	}
	\maketitle
	
	\noindent{\bf Abstract}
	
	\textit{
	ClinicalTrials.gov (CT.gov) is the largest publicly accessible registry of clinical studies, yet its registry-oriented architecture and heterogeneous adverse event (AE) terminology limit systematic pharmacovigilance (PV) analytics. AEs are typically recorded as investigator-reported text rather than standardized identifiers, requiring manual reconciliation to identify coherent safety concepts. We present the ClinicalTrials.gov Transformation Database (CTG-DB), an open-source pipeline that ingests the complete CT.gov XML archive and produces a relational database aligned to standardized AE terminology using the Medical Dictionary for Regulatory Activities (MedDRA). CTG-DB preserves arm-level denominators, represents placebo and comparator arms, and normalizes AE terminology using deterministic exact and fuzzy matching to ensure transparent and reproducible mappings. This framework enables concept-level retrieval and cross-trial aggregation for scalable placebo-referenced safety analyses and integration of clinical trial evidence into downstream PV signal detection.}

	\section{Introduction}

		Drug safety, or pharmacovigilance (PV), involves the systematic evaluation of medicines and vaccines to characterize and monitor their benefit-risk balance \cite{beninger2018pharmacovigilance}. Although post-marketing surveillance remains foundational for signal detection, clinical trial data—spanning early development through post-marketing studies—represent a complementary and comparatively underutilized evidentiary layer for lifecycle safety evaluation. 
		
		ClinicalTrials.gov (CT.gov) is the largest publicly accessible registry of interventional and observational studies, containing structured records for more than 569{,}000 trials across therapeutic areas and phases of development, including both pre-approval and post-marketing studies. The registry was created following the U.S. Food and Drug Administration Modernization Act (FDAMA) of 1997 and launched in 2000 to improve public access to information about clinical studies and promote transparency in the clinical research enterprise \cite{zarin2011results,califf2025importance}. Subsequent policy initiatives—including requirements from the International Committee of Medical Journal Editors (ICMJE) and the U.S. Food and Drug Administration Amendments Act (FDAAA 801)—expanded trial registration and introduced mandatory reporting of summary results and adverse events (AEs) \cite{zarin2017update}.
		
		Despite its scale, CT.gov was designed primarily as a regulatory transparency and study registration platform rather than a PV analytics environment. The native interface supports only keyword-based search and does not enable concept-level queries over AEs. For example, searching for a concept such as “nausea” returns trials where the term appears anywhere in the record—including trials studying nausea itself—without distinguishing trials that report nausea as an AE. In addition, AEs reported in trial results may appear under multiple lexical variants (e.g., ``nausea,'' ``vomiting,'' or ``nausea and vomiting''), and the interface provides no mechanism to reconcile related expressions into clinically coherent concepts. 
		
		Although CT.gov records sometimes indicate that AEs were coded using controlled vocabularies such as the Medical Dictionary for Regulatory Activities (MedDRA), the registry typically exposes only the textual label rather than the underlying coded identifier. AE terminology therefore varies across studies and time periods, reflecting differences in reporting practices and historical vocabularies (e.g., the Common Terminology Criteria for Adverse Events (CTCAE) in earlier trials). Consequently, identifying trials associated with a given AE concept requires anticipating multiple synonymous or related expressions and reconciling them manually during analysis. 
		
		To support computational access, the Aggregate Analysis of ClinicalTrials.gov (AACT) database provides a relational export of CT.gov content \cite{tasneem2012database}. However, AACT preserves registry-native data representations and does not perform terminology normalization or semantic harmonization \cite{AACTGuide}. AE terms are therefore stored largely as investigator-entered text strings, and AACT does not provide mappings to MedDRA identifiers, Standardized MedDRA Queries (SMQs), or other biomedical ontologies. Even when working directly with the database, researchers must perform substantial manual curation and terminology reconciliation to support reproducible safety analyses. 
		
		These characteristics limit systematic PV research using CT.gov data. AEs appear as heterogeneous text expressions with spelling variation and grading annotations; demographic attributes are often reported only at the arm level; and intervention fields encode heterogeneous naming conventions without linkage to standardized drug vocabularies. Recent evaluations of public clinical trial safety reporting show that even when MedDRA-coded AE tables are available, key analytic elements needed for PV are often missing or non-reproducible: exposure-adjusted rates are rarely provided, and the methods used to define events of interest (EOIs)—including which MedDRA terms were grouped or how adjudication was performed—are frequently unspecified or inconsistent across sources \cite{hendrickson2025publicly}. As a result, cross-trial safety contextualization often devolves into manual, expert-driven “detective work,” limiting reproducibility and scalability for lifecycle safety evaluation. 
		
		Collectively, these limitations mean that answering even basic PV questions using CT.gov data requires substantial manual terminology reconciliation and dataset restructuring, introducing process inefficiencies and increasing the risk of inconsistent or error-prone analyses.
		
		In prior work (PVLens) \cite{painter2025pvlens}, we demonstrated that ontology-aligned \mbox{normalization} of regulatory corpora enables reproducible safety analyses. Extending this approach upstream, we present CTG-DB, an open-source transformation pipeline that ingests the complete CT.gov XML archive and produces a safety-oriented relational database designed for deployment in modern relational database management systems. CTG-DB incorporates terminology normalization to reconcile heterogeneous AE expressions by aligning reported terms to MedDRA concepts using exact string matching and configurable fuzzy matching, and subsequently linking normalized terms to Unified Medical Language System (UMLS) concept identifiers to enable synonym-aware retrieval and future ontology-based expansion \cite{bodenreider2004unified,painter2010toward,merrill2008meddra}. By \mbox{preserving} arm-level denominators, explicitly modeling placebo and active treatment arms, and normalizing reported AEs to standardized biomedical vocabularies, CTG-DB renders CT.gov data analytically interoperable with PV workflows.

	\section{Methods}

		CTG-DB implements a transformation pipeline that converts the hierarchical XML representation of CT.gov into a relational, ontology-aligned, safety-oriented data model. The goal is not only structural conversion but also representational normalization: preserving analytic denominators, harmonizing demographic attributes, and aligning AE terminology with MedDRA to support reproducible PV workflows. 

		Unlike AACT, which largely mirrors registry-native representations of CT.gov data, CTG-DB performs terminology normalization and semantic alignment of AEs to MedDRA and UMLS concepts, enabling concept-level safety queries across trials.

		\subsection{Data source and acquisition}
		
		CTG-DB ingests the complete CT.gov XML archive. For the implementation described here, the archive was downloaded on February 10, 2026. The archive (approximately 17\,GB) expanded to 569{,}589 individual XML files, each corresponding to a registered study. Study inclusion criteria followed prior registry analyses \cite{tasneem2012database}. Studies were excluded if results were withheld, eligibility criteria were absent, or no conditions were specified. These filters removed 25,274 studies (4.4\%), yielding 544{,}315 studies for ingestion. The resulting dataset spans randomized trials, non-randomized interventional studies, and post-marketing investigations registered in CT.gov. CTG-DB preserves arm-level representations where available but does not restrict inclusion to randomized controlled trials.  The continued expansion of the registry underscores the importance of scalable, terminology-aligned transformation pipelines.

		\subsection{XML parsing and intermediate representations}
		
		The native CT.gov XML schema is deeply hierarchical and variably populated. To enable deterministic extraction, the XML parser was implemented in Java using Jakarta XML Binding (JAXB) \cite{fialli2003java}. The CT.gov XML Schema Definition (XSD) was reverse-engineered into strongly typed Java classes, allowing structured de-marshalling of XML elements into typed objects.
		
		This approach reduces the risk of silent field omission when schema elements evolve and enforces explicit handling of optional and repeating elements. Parsed objects were subsequently serialized into normalized, table-oriented CSV files designed for bulk ingestion into relational database systems.
		
		\subsection{Demographic extraction and harmonization}
		
		CTG-DB prioritizes arm-level resolution to support stratified safety analyses and cross-trial comparability. Trial-level eligibility elements—including minimum age, maximum age, and sex eligibility—were extracted and denormalized to facilitate cohort-level filtering.
		
		At the arm level, CTG-DB captures the number of participants who started each arm, sex counts, age summary statistics when available, and reported ethnicity counts. Preserving arm-level denominators enables proportional safety analyses and consistent comparison of AE frequencies across heterogeneous trials.
		
		Ethnicity reporting in CT.gov is heterogeneous and inconsistently populated. Distinct ethnicity strings were therefore enumerated across the archive and mapped to a reduced set of harmonized categories when feasible. This harmonization serves as a normalization layer to improve analytic comparability rather than to impose semantic completeness. Missing values are preserved rather than imputed to avoid introducing artificial structure into the source data. CTG-DB additionally derives whether healthy volunteers were included in a study, as this attribute materially affects interpretation of AE frequencies and population comparability \cite{kadam2016challenges,lim2017simulating}. This indicator supports analytic stratification when comparing event rates across heterogeneous study populations.
		
		\subsection{Adverse event terminology normalization to MedDRA}
		
		Although CT.gov provides MeSH mappings for some AEs, documentation indicates that these mappings are incomplete and may not reflect PV-specific semantics. Because regulatory and post-marketing workflows primarily operate on MedDRA \cite{merrill2008meddra,merrill2008construction,ly2018evaluation}, CTG-DB adopts MedDRA as the canonical terminology for AE and condition terms.
		
		Normalization uses a two-stage alignment strategy:
		
		\textbf{Stage 1: Deterministic lexical matching.} Reported AE strings are compared using exact, case-insensitive lexical matching to MedDRA Preferred Terms (PTs) and Lower-Level Terms (LLTs). This step captures directly codable terms while preserving transparent mappings.
		
		\textbf{Stage 2: Bigram-based fuzzy alignment.} Terms failing exact matching are processed using bigram similarity to identify likely lexical variants, following principles described for terminology inference in unstructured vocabularies \cite{painter2010toward}. This step reduces fragmentation caused by misspellings, punctuation artifacts, minor orthographic variants, and appended grading annotations (e.g., ``G1'', ``G3'') while maintaining traceability to normalized MedDRA codes. 

		Across the processed archive, 138{,}032 unique reported AE strings were observed. Of these, 18.41\% (25{,}407) matched MedDRA via exact lexical alignment and an additional 40.82\% (56{,}340) via bigram-based fuzzy matching, yielding 59.22\% total coverage at the unique-string level (Table~\ref{tab:ae_coverage}).
		
		Because PV relevance is driven by event burden rather than vocabulary diversity, coverage was also evaluated by weighting counts by arm-level participants. When weighted by participants affected (27{,}469{,}697 total AE counts), 88.76\% of AE burden mapped via exact alignment and 6.25\% via fuzzy matching, yielding 95.01\% total mapped burden (Table~\ref{tab:ae_coverage}). Coverage is therefore reported from both lexical (unique-string) and analytic (burden-weighted) perspectives.

		\begin{table}[t]
			\centering
			\caption{Coverage of CT.gov adverse event terms mapped to MedDRA, reported at the unique-string level and weighted by participants affected.}
			\label{tab:ae_coverage}
			\begin{tabular}{lrrrr}
				\toprule
				& \multicolumn{2}{c}{Unique reported AE strings} & \multicolumn{2}{c}{Weighted by participants affected} \\
				\cmidrule(lr){2-3} \cmidrule(lr){4-5}
				Mapping category & Count & Percentage (\%) & Participants affected & Percentage (\%) \\
				\midrule
				Exact lexical match & 25{,}407 & 18.41 & 24{,}382{,}112 & 88.76 \\
				Fuzzy (bigram) match & 56{,}340 & 40.82 & 1{,}717{,}553 & 6.25 \\
				Unmapped & 56{,}285 & 40.78 & 1{,}370{,}032 & 4.99 \\
				\midrule
				Total mapped & 81{,}747 & 59.22 & 26{,}099{,}665 & 95.01 \\
				Total  & 138{,}032 & 100.00 & 27{,}469{,}697 & 100.00 \\
				\bottomrule
			\end{tabular}
		\end{table}

		An analogous evaluation was performed for condition terms reported in CT.gov study records, where 54.27\% of unique condition strings mapped to MedDRA (10.51\% exact; 43.77\% fuzzy). Representative lexical variants (e.g., ``Nausea?'', ``Nausea G1'', or misspellings) were consolidated to a single MedDRA PT during normalization. This hybrid approach reduces lexical noise while preserving regulatory-aligned terminology and analytic denominators. The normalization layer also enables transparent, reusable definitions of events of interest (EOIs) through aggregation across related MedDRA terms, addressing limitations commonly observed in public clinical trial safety reporting \cite{hendrickson2025publicly}.

		\subsection{Relational schema design and implementation}
		
		The CTG-DB schema centers on the \texttt{clinical\_trial} table, which stores identifying metadata (e.g., NCT identifier, titles, summaries, registry URLs) and links to dimension tables for study status, phase, type, geography, conditions, and interventions. Population structure is modeled explicitly at the arm level through the \texttt{ct\_arms} table, preserving denominators and comparator structure. In CT.gov, these groups may represent randomized treatment arms or analytic cohorts in observational studies, reflecting the registry's standardized reporting structure.
		
		Arm records link to AE count tables distinguishing serious and non-serious outcomes, consistent with CT.gov reporting conventions. Because CT.gov provides only aggregated outcomes and no patient-level identifiers, CTG-DB stores event counts at the arm level and contains no person-level data, aligning with ethical and reporting constraints \cite{deichmann2016bioethics}.
		
\begin{figure}[!ht]
			\centering
			\includegraphics[width=0.8\textwidth]{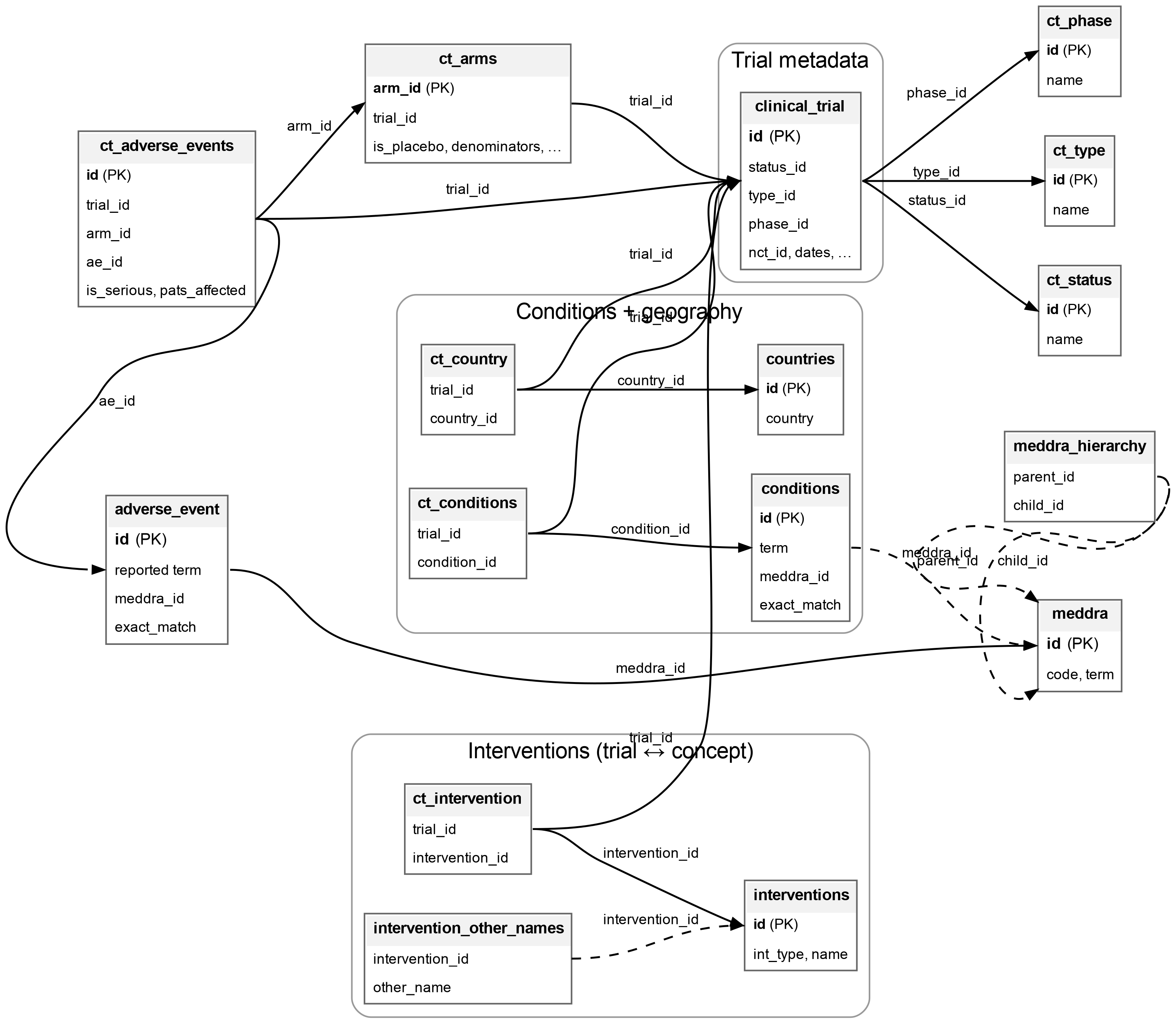}
			\caption{
				Conceptual overview of the CTG-DB relational schema showing the primary tables and relationships linking trial metadata, arm-level structures, and adverse event normalization to MedDRA. Arm-level denominators and comparator structure are preserved to support placebo-referenced cross-trial aggregation. Only principal tables and relationships are shown. 			
			}
			\label{fig:ctgdb_schema}
		\end{figure}

		The overall relational architecture is illustrated in Figure~\ref{fig:ctgdb_schema}. The schema integrates reported AE terms with MedDRA through a normalization layer and preserves MedDRA hierarchy relationships to support class-level aggregation and ontology-aware analyses.
		
		The schema preserves analytic denominators (participants starting each arm), enabling proportional analyses, placebo-referenced screening, and cross-trial aggregation within therapeutic classes.
		
		\subsection{Database creation and ingestion workflow}
		
		Database creation and ingestion were orchestrated in Python. The pipeline generates data definition language (DDL) statements directly from the schema specification and loads normalized intermediate CSV outputs into their corresponding tables using bulk-loading utilities, with \texttt{pandas} used for controlled type handling and validation.
		
		End-to-end ingestion is instrumented with structured logging to ensure reproducibility and auditability. Although early iterations used SQLite for rapid prototyping, CTG-DB is designed for deployment in modern relational database management systems. The logical schema is database-agnostic, and migration scripts provided in the project repository support deployment across systems such as MySQL and PostgreSQL while maintaining portability and version consistency.

	\section{Case study: placebo-referenced cross-trial screening}
			
			To illustrate the analytic affordances of CTG-DB, we conducted a proof-of-concept evaluation using publicly available CT.gov arm-level data within an anonymized therapeutic class. The objective was methodological demonstration rather than confirmatory inference.
			
		\subsection{Case study implementation}
			
			Arm-level data were extracted from the CTG-DB relational schema and aggregated at both the study-arm and product levels.
			
			Arm-level event proportions were computed directly from CT.gov-reported counts and denominators. Placebo arms were pooled across eligible studies to derive empirical reference distributions. Product-level summaries were calculated by aggregating across trials, with restriction to Phase 3 and Phase 4 studies to enhance clinical and developmental comparability.
			
			Comparative metrics, including odds ratios (ORs), were computed using aggregated counts with placebo as the reference category. These calculations illustrate how CTG-DB enables rapid assembly of cross-trial analytic datasets while preserving transparent linkage to the underlying registry data.			

		\subsection{Event definition and MedDRA grouping}
			
			Reported AEs were normalized to MedDRA (v24.1) PTs using the two-stage normalization strategy described above. To define a clinically coherent analytic construct, multiple related MedDRA PTs representing gastrointestinal hemorrhage (e.g., upper gastrointestinal hemorrhage, lower gastrointestinal hemorrhage, gastrointestinal hemorrhage) were selected and aggregated into a single event grouping. This illustrates a core limitation of registry-native reporting: clinically meaningful safety concepts typically span multiple related terms, yet public sources often do not specify standardized term groupings or provide concept-level frequencies in a reusable way \cite{hendrickson2025publicly}.
			
			This grouping reflects a PV-relevant conceptual cluster rather than reliance on a single PT, thereby aligning event definition with regulatory interpretability while preserving transparent mapping to underlying MedDRA codes.
			
		\subsection{Arm-level event proportion}
			
			For each study arm, the AE proportion was computed as:
			
			\[
			P_{\mathrm{arm}} = \frac{n_{\mathrm{AE}}}{n_{\mathrm{started}}},
			\]
			
			where $n_{\mathrm{AE}}$ denotes the number of participants in the arm reported with the mapped AE group and $n_{\mathrm{started}}$ denotes the number of participants who initiated the arm, consistent with CT.gov reporting conventions.
			
			Preservation of the arm-level denominator is central to this analytic design, enabling proportional comparisons across trials without imputation of participant counts.
			
		\subsection{Placebo aggregation and reference thresholds}
			
			Placebo arms across eligible trials were aggregated to derive an empirical reference distribution of $P_{\mathrm{arm}}$. From this distribution, the 75th percentile and the maximum observed placebo proportion were computed and used as heuristic screening thresholds. The 75th percentile was selected as a pragmatic upper-quartile screening heuristic rather than as a formal inferential threshold.
			
			This placebo-referenced framework provides contextualization for cross-trial heterogeneity and enables identification of investigational arms with event frequencies exceeding empirically observed placebo variation. The thresholds are descriptive rather than inferential and are intended to prioritize arms for further evaluation.

		\subsection{Comparative metrics}
			
			To support cross-product screening, aggregated event proportions were computed within each investigational product across trials, with restriction to Phase 3 and Phase 4 studies to reduce heterogeneity introduced by early-phase dose-finding designs.
			
			Using the aggregated placebo proportion as the reference, odds ratios (ORs) were computed for product-versus-placebo comparisons. Where applicable, head-to-head product comparisons were also calculated. ORs were additionally computed at the study-arm level relative to the pooled placebo proportion to prioritize individual arms for review.
			
			These comparative metrics are intended for signal screening and hypothesis generation rather than for confirmatory statistical testing. Because arms are aggregated across studies, the original randomization structure within individual trials is not preserved in these comparisons. Approaches that explicitly condition on the study effect (e.g., stratified or conditional logistic regression) could provide a design-aware extension in future analyses. No multiplicity adjustment was applied, as the objective was exploratory prioritization rather than confirmatory hypothesis testing.

		\subsection{Visualization of arm-level screening results}
			
			\begin{figure}[!ht]
				\centering
				\includegraphics[width=0.9\textwidth]{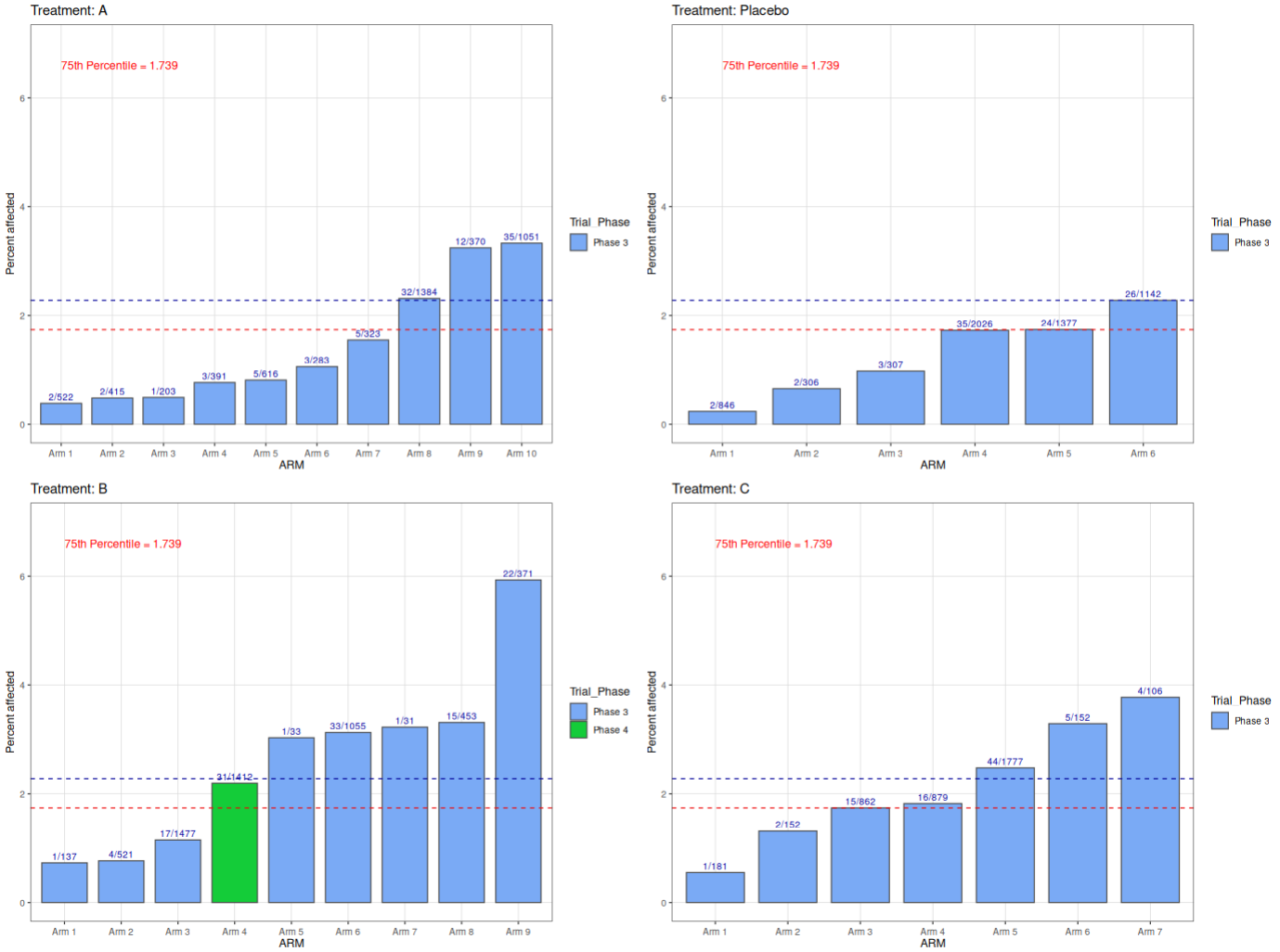}
				\caption{
					Arm-level adverse event proportions across anonymized investigational products (A–C) and placebo. Each bar represents the proportion of participants experiencing the MedDRA-normalized event within a single clinical trial arm. The red dashed line indicates the 75th percentile of aggregated placebo proportions and the blue dashed line indicates the maximum observed placebo proportion. Trial phase is color-coded.
				}
				\label{fig:ctgdb_demo}
			\end{figure}
			
			Figure \ref{fig:ctgdb_demo} displays arm-level AE proportions across investigational products and aggregated placebo arms. Several investigational arms exceed the placebo 75th percentile threshold, and a subset exceed the maximum observed placebo proportion. Elevated proportions are not uniformly distributed across development phases, underscoring the importance of arm-level granularity and phase-aware interpretation when aggregating clinical trial safety data.
			
			Because CT.gov provides aggregated arm-level counts without patient-level linkage, these placebo-referenced comparisons do not adjust for between-study heterogeneity, follow-up duration, multiplicity, or covariate imbalance. Although individual clinical trials are designed so that treatment arms are internally comparable through randomization, this comparability is not preserved when arm-level data are aggregated across studies. Consequently, the resulting analyses should be interpreted with caution as screening-level signals intended to prioritize arms or products for further evaluation rather than as estimates of causal treatment effects. Nonetheless, this framework demonstrates how CTG-DB enables rapid identification of arms and products warranting deeper investigation within a reproducible, ontology-aligned analytic environment.

		\section{Discussion}
			
			The findings of this work should be interpreted in the context of structural limitations in CT.gov data. Although the registry contains extensive safety information across hundreds of thousands of studies, AEs are typically represented as heterogeneous text expressions rather than standardized coded identifiers. Consequently, even simple safety questions—such as identifying trials reporting a given clinical event—require substantial manual terminology reconciliation when using the CT.gov interface or relational exports such as AACT. CTG-DB addresses this limitation by transforming registry-native data into a concept-aligned relational structure designed for reproducible PV analysis. 
			
			CTG-DB is not simply a relational re-expression of CT.gov but a representational transformation designed to support established PV workflows. Its central contribution is converting a registry-native XML archive into an ontology-aligned, denominator-preserving analytic substrate for cross-trial safety evaluation. Public trial safety evidence is often difficult to reuse analytically because higher-order constructs such as events of interest (EOIs) are inconsistently defined and exposure-adjusted rates are rarely available, limiting reproducibility and comparability \cite{hendrickson2025publicly}. CTG-DB addresses these limitations by preserving arm-level denominators and enabling transparent MedDRA-based aggregation for concept-level screening. 
			
			MedDRA normalization reduces lexical fragmentation in AE reporting and renders CT.gov safety data interoperable with regulatory and post-marketing systems. Explicit arm-level modeling preserves denominators and comparator structure, enabling proportional aggregation across heterogeneous trials. 
		
			Across the CT.gov archive, 59.2\% of unique AE strings aligned to MedDRA concepts (18.4\% exact; 40.8\% fuzzy). When weighted by affected participants, coverage increased to 95.0\%, indicating that the normalization strategy captures most clinically relevant safety events while leaving primarily rare lexical variants unmapped. Together, these design choices expose analytic capabilities implicit in the registry but not directly accessible without systematic transformation. 
			
			Although the MedDRA alignment strategy improves terminology consistency, automated matching cannot fully replace manual curation. Some AE expressions remain ambiguous or map imperfectly to standardized concepts, particularly when investigator-entered descriptions contain composite phrases or context-dependent qualifiers. 
			
			Relative to AACT \cite{tasneem2012database}, CTG-DB is distinguished by three design principles: (i) systematic MedDRA normalization of AEs; (ii) preservation and harmonization of arm-level attributes to support stratified and phase-restricted analyses; and (iii) explicit support for placebo-referenced aggregation within therapeutic classes. These design choices reflect core PV requirements, including standardized terminology, traceable ontology mappings, and explicit denominators for defensible cross-study comparisons \cite{merrill2008meddra,seagrave2017adverse,hendrickson2025publicly}.
		
			CTG-DB does not attempt to reconstruct patient-level trajectories or infer individual exposure histories. Instead, it preserves the aggregate structure of CT.gov reporting while enhancing interpretability through ontological alignment and reproducible aggregation logic. Accordingly, the system is intended to support screening-level evaluation and lifecycle-oriented safety synthesis rather than confirmatory causal inference.

			\subsection{Limitations and future work}

			A primary limitation involves intervention normalization. CT.gov intervention fields encode combination regimens, investigational codes, and comparators within semi-structured text, often referencing multiple products without standardized naming. These ambiguities constrain class-level aggregation and cross-source linkage. 
			
			Future work will focus on mapping interventions to standardized drug concepts (e.g., RxNorm and active moieties), resolving multi-product arms, and formalizing comparator identification rules. Additional enhancements include strengthened denominator validation and uncertainty-aware aggregation strategies to account for cross-trial heterogeneity. 
			
			CT.gov reporting practices also vary across studies and time periods. Differences in reporting thresholds, grading conventions, and arm-level denominators may influence cross-trial comparisons and should be considered when interpreting aggregated safety signals. 
			
			Cross-trial aggregation of arm-level AE counts also abstracts certain design characteristics of the original trials. Differences in randomization ratios, allocation schemes, and comparator structures are not preserved when aggregating event frequencies across studies. Consequently, proportional comparisons derived from CTG-DB should be interpreted as screening-level evidence rather than estimates of causal treatment effects. Future work may incorporate trial-weighting strategies or design-aware aggregation approaches to better account for allocation ratios and study structure when synthesizing safety signals across heterogeneous trials.

			\subsection{Extension: clinical development priors for Bayesian safety signal detection}
			
			Beyond within-source placebo-referenced screening, CTG-DB enables integration of clinical development evidence into post-marketing PV signal detection. In a separate evaluation using FDA Adverse Event Reporting System (FAERS) data (2015–2019), CT.gov-derived disproportionality estimates were incorporated as robustified meta-analytic predictive priors within a Bayesian dynamic borrowing framework, producing a clinical-trial informed information component (CTIC). 
			
			Although overlap between CT.gov and spontaneous reporting evidence was limited (approximately 6\% of FAERS product–event pairs had corresponding CT.gov data), incorporation of clinical development evidence produced modest but consistent performance improvements relative to traditional disproportionality methods. 
			
			This application demonstrates that CTG-DB allows clinical trial evidence to function as structured prior information within Bayesian safety signal detection frameworks, extending its utility from placebo-referenced screening to lifecycle-wide evidence synthesis.
	
	\section{Conclusion}
		
		CT.gov is the largest publicly accessible repository of clinical trial evidence, yet its XML structure and heterogeneous terminology limit systematic safety analysis. Recent reviews of public clinical trial safety sources emphasize that missing exposure-adjusted rates and unclear event-of-interest (EOI) definitions limit reuse of these data for PV decision-making and signal contextualization \cite{hendrickson2025publicly}. CTG-DB addresses this gap by transforming the archive into an ontology-aligned, denominator-preserving relational database designed for PV workflows.

		By preserving arm-level denominators, explicitly modeling placebo and comparator structure, and normalizing AEs to MedDRA, CTG-DB enables cross-trial proportional aggregation and placebo-referenced screening not directly accessible from the registry alone.
		 
		Beyond within-source screening, CTG-DB supports incorporation of clinical development evidence into Bayesian safety signal detection frameworks. Released as open-source software with portable MySQL and PostgreSQL schemas, CTG-DB lowers barriers to transparent, reproducible safety analytics and strengthens evidentiary continuity across the product lifecycle. 		
	
	\section{Declarations} 
		GSK covered all costs associated with the conduct of the study and the development of the manuscript and the decision to publish the manuscript. J.P., F.H. and A.B. are employed by GSK and hold financial equities. This manuscript has not been submitted to, nor is under review at, another journal or other publishing venue. The authors have no competing interests to declare that are relevant to the content of this article.\\
		
		\textbf{Author Contributions} JLP, FH, and AB contributed to the study concept, data acquisition, data analysis, and data interpretation.\\

		\textbf{Data availability} All code and documentation to support this work are available at: \\ \url{https://github.com/jlpainter/ctgdb} 
	
	\bibliographystyle{unsrturl}
	\bibliography{ctgdb.bib}

@article{beninger2018pharmacovigilance,
	title        = {{Pharmacovigilance: an overview}},
	author       = {Beninger, Paul},
	year         = {2018},
	journal      = {Clinical Therapeutics},
	volume       = {40},
	number       = {12},
	pages        = {1991--2004},
	doi          = {10.1016/j.clinthera.2018.07.012}
}

@article{zarin2011results,
	title        = {{The ClinicalTrials.gov results database---update and key issues}},
	author       = {Zarin, Deborah A. and Tse, Tony and Williams, Rebecca J. and Califf, Robert M. and Ide, Nicholas C.},
	year         = {2011},
	journal      = {New England Journal of Medicine},
	volume       = {364},
	number       = {9},
	pages        = {852--860},
	doi          = {10.1056/NEJMsa1012065}
}

@article{califf2025importance,
	title        = {{The importance of ClinicalTrials.gov in informing trial design, conduct, and results}},
	author       = {Califf, Robert M. and Cutler, Tracy L. and Marston, Hilary D. and Meeker-O'Connell, Ann},
	year         = {2025},
	journal      = {Journal of Clinical and Translational Science},
	volume       = {9},
	number       = {1},
	pages        = {e42},
	doi          = {10.1017/cts.2025.9}
}

@article{zarin2017update,
	title        = {{Update on trial registration 11 years after the ICMJE policy was established}},
	author       = {Zarin, Deborah A. and Tse, Tony and Williams, Rebecca J. and Rajakannan, Thiyagu},
	year         = {2017},
	journal      = {New England Journal of Medicine},
	volume       = {376},
	number       = {4},
	pages        = {383--391},
	doi          = {10.1056/NEJMsr1601330}
}

@article{tasneem2012database,
	title        = {{The database for aggregate analysis of ClinicalTrials.gov (AACT) and subsequent regrouping by clinical specialty}},
	author       = {Tasneem, Asba and Aberle, Laura and Ananth, Hari and Chakraborty, Swati and Chiswell, Karen and McCourt, Brian J. and Pietrobon, Ricardo},
	year         = {2012},
	journal      = {PLoS One},
	volume       = {7},
	number       = {3},
	pages        = {e33677},
	doi          = {10.1371/journal.pone.0033677}
}

@misc{AACTGuide,
	title        = {{Points to Consider When Using the AACT Database}},
	author       = {{Clinical Trials Transformation Initiative (CTTI)}},
	year         = {2024},
	url          = {https://aact.ctti-clinicaltrials.org/points_to_consider},
	note         = {Accessed Feb 28, 2026}
}

@article{hendrickson2025publicly,
	title        = {{Publicly Available Clinical Trial Safety Data: Review and a Call for Standardization and Improved Reporting Practices}},
	author       = {Hendrickson, Barbara A. and McShea, Cynthia and Hammad, Tarek A. and Meer, Jaishri and Zhang, Wei Michelle and Whalen, Edward and Talbot, Susan and Sinvhal, Ranjeeta and Panico, Maria Beatrice and Lin, Li-An and others},
	year         = {2025},
	journal      = {Pharmaceutical Medicine},
	pages        = {1--17},
	doi          = {10.1007/s40290-025-00596-5}
}

@misc{painter2025pvlens,
	title        = {{PVLens: Enhancing pharmacovigilance through automated label extraction}},
	author       = {Painter, Jeffery L. and Powell, Gregory E. and Bate, Andrew},
	year         = {2025},
	month        = nov,
	doi          = {10.48550/arXiv.2503.20639},
	howpublished = {Preprint. Presented at the AMIA Annual Symposium}
}

@article{bodenreider2004unified,
	title        = {{The unified medical language system (UMLS): integrating biomedical terminology}},
	author       = {Bodenreider, Olivier},
	year         = {2004},
	journal      = {Nucleic Acids Research},
	volume       = {32},
	number       = {suppl\_1},
	pages        = {D267--D270},
	doi          = {10.1093/nar/gkh061}
}

@inproceedings{painter2010toward,
	title        = {{Toward automating an inference model on unstructured terminologies: OXMIS case study}},
	author       = {Painter, Jeffery L.},
	year         = {2010},
	booktitle    = {Advances in Computational Biology},
	pages        = {645--651},
	publisher    = {Springer},
	doi          = {10.1007/978-1-4419-5913-3_71}
}

@inproceedings{merrill2008meddra,
	title        = {{The MedDRA Paradox}},
	author       = {Merrill, Gary H.},
	year         = {2008},
	booktitle    = {AMIA Annual Symposium Proceedings},
	volume       = {2008},
	pages        = {470},
	url          = {https://pubmed.ncbi.nlm.nih.gov/18998828/},
	pmid         = {18998828}
}

@techreport{fialli2003java,
	title        = {{The Java architecture for XML binding (JAXB)}},
	author       = {Fialli, Joseph and Vajjhala, Sekhar},
	year         = {2003},
	institution  = {Java Community Process},
	type         = {JSR Specification},
	number       = {JSR 31},
	url          = {https://jcp.org/en/jsr/detail?id=31}
}

@article{kadam2016challenges,
	title        = {{Challenges in recruitment and retention of clinical trial subjects}},
	author       = {Kadam, Rashmi Ashish and Borde, Sanghratna Umakant and Madas, Sapna Amol and Salvi, Sundeep Santosh and Limaye, Sneha Saurabh},
	year         = {2016},
	journal      = {Perspectives in Clinical Research},
	volume       = {7},
	number       = {3},
	pages        = {137},
	doi          = {10.4103/2229-3485.184820}
}

@article{lim2017simulating,
	title        = {{Simulating clinical trial visits yields patient insights into study design and recruitment}},
	author       = {Lim, S. Sam and Kivitz, Alan J. and McKinnell, Doug and Pierson, M. Edward and O'Brien, Faye S.},
	year         = {2017},
	journal      = {Patient Preference and Adherence},
	volume       = {11},
	pages        = {1295},
	doi          = {10.2147/PPA.S137416}
}

@inproceedings{merrill2008construction,
	title        = {{Construction and annotation of a UMLS/SNOMED-based drug ontology for observational pharmacovigilance}},
	author       = {Merrill, Gary H. and Ryan, Patrick B. and Painter, Jeffery L.},
	year         = {2008},
	booktitle    = {Proceedings of the Intelligent Data Analysis in Biomedicine and Pharmacology (IDAMAP) Workshop},
	address      = {Washington, DC, USA},
	note         = {Presented at IDAMAP, held in conjunction with the AMIA Annual Symposium}
}

@article{ly2018evaluation,
	title        = {{Evaluation of Natural Language Processing (NLP) systems to annotate drug product labeling with MedDRA terminology}},
	author       = {Ly, Thomas and Pamer, Carol and Dang, Oanh and Brajovic, Sonja and Haider, Shahrukh and Botsis, Taxiarchis and Milward, David and Winter, Andrew and Lu, Susan and Ball, Robert},
	year         = {2018},
	journal      = {Journal of Biomedical Informatics},
	volume       = {83},
	pages        = {73--86},
	doi          = {10.1016/j.jbi.2018.05.019}
}

@article{deichmann2016bioethics,
	title        = {{Bioethics in practice: Considerations for stopping a clinical trial early}},
	author       = {Deichmann, Richard E. and Krousel-Wood, Marie and Breault, Joseph},
	year         = {2016},
	journal      = {Ochsner Journal},
	volume       = {16},
	number       = {3},
	pages        = {197--198},
	url          = {https://pubmed.ncbi.nlm.nih.gov/27660563/}
}

@article{seagrave2017adverse,
	title        = {{Adverse drug reactions}},
	author       = {Seagrave, Zachary and Bamba, Sonya},
	year         = {2017},
	journal      = {Disease-a-Month},
	volume       = {63},
	number       = {2},
	pages        = {49--53},
	doi          = {10.1016/j.disamonth.2016.09.006}
}
\end{document}